\documentclass[lettersize,journal]{IEEEtran}
\usepackage{amsmath,amsfonts}
\usepackage{algorithmic}
\usepackage{algorithm}
\usepackage{array}
\usepackage{textcomp}
\usepackage{stfloats}
\usepackage{url}
\usepackage{verbatim}
\usepackage{graphicx}
\usepackage{cite}
\usepackage{amssymb}
\usepackage[normalem]{ulem}  
\usepackage{booktabs}
\usepackage{makecell} 
\usepackage{threeparttable}
\usepackage{multirow}

\usepackage{graphicx}
\usepackage[caption=false,font=footnotesize]{subfig}

\usepackage{enumitem}
\usepackage{siunitx}

\usepackage{cite}

\usepackage{color}
\usepackage{kotex}

\begin{document}

\title{GAIT: Legged Robot Proprioceptive State Estimation with Attention over Inertial–Leg Tokens}

\author{
Young-Rang Seo$^{1}$, Hajun Kim$^{1}$, Sangmin Kim$^{1}$, Dongyun Kang$^{1}$ and Hae-Won Park$^{1}$
\thanks{$^{1}$Korea Advanced Institute of Science and Technology, Yuseong-gu, Daejeon 34141, Republic of Korea. haewonpark@kaist.ac.kr
}
}



\maketitle

\begin{abstract}
In this paper, we propose a method that applies Inertial-Leg (IL) tokenization to an attention-based network for proprioceptive state estimation in legged robots. 
Unlike existing learning-based state estimators that concatenate all sensor measurements into a single flat vector, the proposed architecture represents inertial measurements and leg-wise measurements as individual tokens and uses an attention mechanism to learn the relative importance of each measurement.
This design allows the network to reweight each measurement according to the current contact condition, reflecting the fact that the reliability of forward kinematic measurements depends on whether the corresponding foot is in contact. 
Unlike conventional contact-aided estimators, however, the proposed method learns this behavior without relying on an explicit contact estimator or on explicit measurement updates based on a stationary contact assumption.
To validate the proposed method, we conducted experiments on a Unitree Go1 robot, including debris terrain not modeled in simulation and gait patterns not seen during training.
Experimental results show that the proposed method achieves better estimation performance than existing learning-based state estimators under unseen gait patterns and also improves performance over contact-aided model-based methods.
\end{abstract}


\section{INTRODUCTION}


Recently, legged robots have shown remarkable mobility across challenging terrains and dynamic tasks~\cite{lee2020learning}.
For such locomotion tasks, accurate and robust state estimation is essential for feedback control and motion planning~\cite{barfoot2024state}. 
Existing state estimators often fuse proprioceptive measurements from IMUs and joint encoders with exteroceptive measurements from cameras or LiDAR~\cite{wisth2019robust, kim2022step, wisth2022vilens}. 
However, exteroceptive sensors often operate at lower update rates and can degrade under sensing conditions such as poor illumination for cameras or limited geometric structure for LiDAR~\cite{lee2024lidar}.
Moreover, since exteroceptive measurements are typically fused with a proprioceptive estimate, a robust proprioceptive foundation is essential even when such sensors are used. 
Therefore, proprioceptive-only state estimation remains important for legged robots that require high-frequency and reliable feedback.


To address this need, proprioceptive-only state estimation for legged robots has been largely developed within contact-aided frameworks. 
These approaches typically integrate IMU data with leg kinematics, assuming that the foot in contact with the ground is stationary~\cite{bloeschStateEstimationLegged2013}.
Within this category, a contact-aided invariant extended Kalman filter  was introduced in ~\cite{hartleyContactaidedInvariantExtended2020}, demonstrating improved convergence and consistency relative to conventional quaternion-based EKFs.

However, these methods can degrade significantly when the stationary contact assumption is violated, such as during slippage.
To mitigate the slippage problem,~\cite{bloesch2013state} proposed an innovation-based outlier rejection scheme within a UKF framework, which filters out kinematic measurements that exceed a Mahalanobis-distance threshold.
More recent studies further improved the robustness to such violations through smoothing and slip rejection mechanisms ~\cite{kimLeggedRobotState2021, yoonInvariantSmootherLegged2024}. Nevertheless, model-based approaches often rely heavily on heuristic parameter tuning and require explicit contact estimators or contact sensors.


\begin{figure}[t]
    \centering
    \includegraphics[width=\columnwidth]{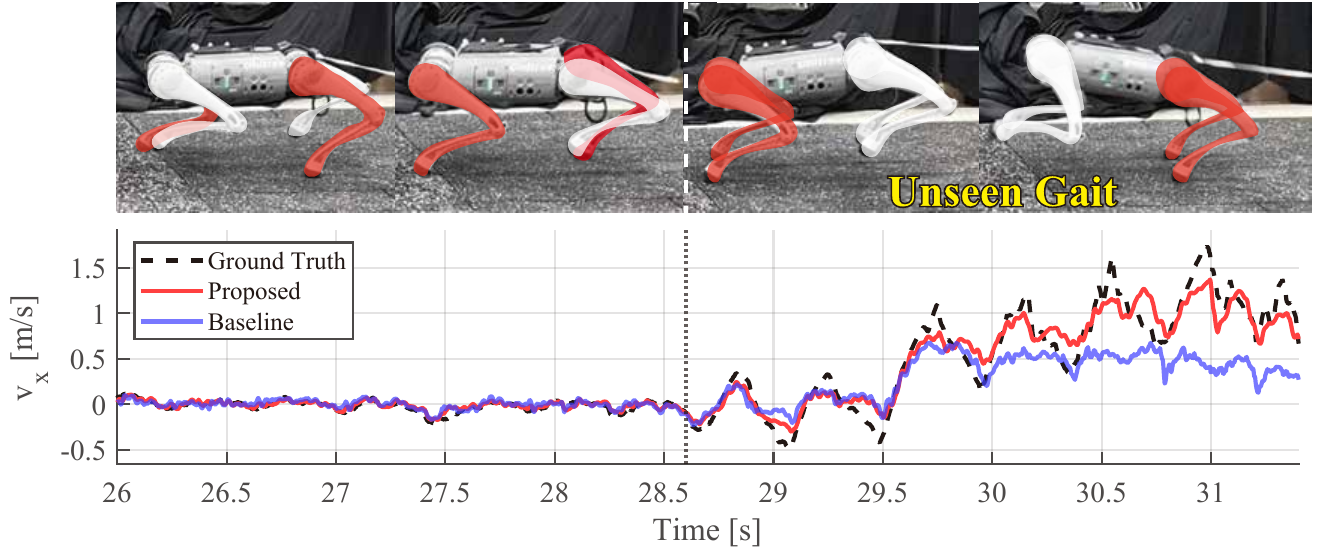}
    \caption{
    An example of the estimated body-frame linear velocity along the x-axis during a gait transition from trot to bound and the corresponding attention weights of the proposed network.
    Although trained only on trot gait, the proposed network shows lower estimation error than the baseline network-based state estimator~\cite{youmLeggedRobotState2024} (blue), even for the unseen bound gait.
    The dashed line marks the gait transition point.}
    \label{fig:figure1}
\end{figure}


In response to these limitations, recent learning-based approaches have incorporated neural networks into proprioceptive state estimation, aiming to improve robustness to slippage without relying solely on heuristic tuning~\cite{lin2022legged, wasserman2024legolas, youmLeggedRobotState2024, lee2026attention}.
In particular, several works~\cite{wasserman2024legolas, youmLeggedRobotState2024, lee2026attention} train learning-based estimators using simulation-generated datasets to reduce the reliance on large-scale real-world locomotion data.

Despite these advancements, since learning-based estimators are typically trained on data generated by a specific locomotion policy, their generalization outside the training distribution remains uncertain.
\IEEEpubidadjcol
Furthermore, changes in locomotion policies, unseen terrains, and sim-to-real gaps can induce state distributions that were not observed during training, which can in turn degrade the performance of the learning-based state estimators. 
Therefore, improving generalization capability to handle such distribution shifts is essential for enhancing the robustness of learning-based state estimators.

One promising solution is to design network architectures that incorporate the structural properties of the target problem, often referred to as inductive biases~\cite{battaglia2018relational, bronstein2021geometric}.
Such biases can help models capture interactions between relevant entities and improve generalization beyond the training distribution.
In robot control, recent studies have exploited body structure as an inductive bias to improve performance and generalization across different embodiments and tasks~\cite{sferrazza2024body, patel2025get}.

In this study, we aim to improve learning-based proprioceptive state estimation for legged robots by incorporating inductive biases into the network architecture. 
The main contributions of this paper can be summarized as follows:
\begin{itemize}
    \item This work proposes an attention-based proprioceptive state estimator whose tokenized architecture provides an inductive bias for representing contact-dependent variations in measurement reliability, improving generalization under unseen contact conditions.
    \item We introduce inertial--leg (IL) tokenization to help the network capture contact-dependent variations in inertial and leg-wise proprioceptive inputs.
    \item Unlike existing model-based proprioceptive-only approaches, the proposed approach can achieve robust state estimation without relying on explicit contact estimation by integrating the predicted body linear velocity as a measurement factor.
    \item We demonstrate on the Unitree Go1 robot that the proposed framework, trained exclusively on trotting data collected in simulation, outperforms prior methods in real-world experiments conducted on unseen terrain and across gaits not included in the training dataset, without additional fine-tuning. 
    
\end{itemize}

\section{System Overview}
\label{sec:system_design}

\begin{figure*}[t]
    \centering
    \includegraphics[width=\textwidth]{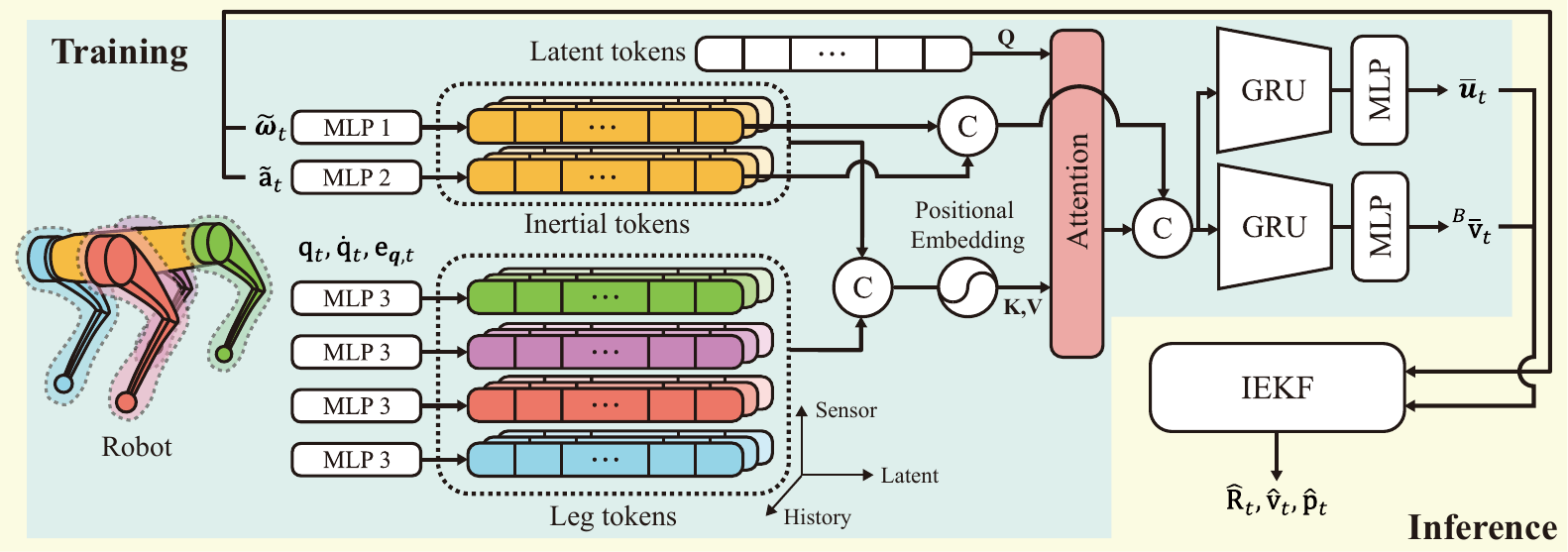}
    \caption{\textbf{Overview of the proposed state estimation framework}: 
    During training, the network estimates the body linear velocity and its associated uncertainty from inertial and leg-wise measurements, with Gaussian noise added only to the inertial measurements.
    During inference, the trained network outputs are incorporated as measurements in the IEKF.
    Each leg token is constructed from the joint position, joint velocity, and joint position tracking error of the corresponding leg, with matching colors indicating the same leg.
    Training data are generated in simulation.}
    \label{fig:framework}
\end{figure*}

The proposed system consists of two main components: (i) an attention-based network utilizing IL tokenization to estimate body linear velocity and its associated uncertainty, and (ii) a filtering module that integrates raw IMU data with the pseudo-measurements predicted by the network. 
The overall framework is illustrated in Fig.~\ref{fig:framework}. 

Existing learning-based estimators typically concatenate IMU and kinematic measurements into a single input vector for pseudo-measurement prediction~\cite{lin2022legged, youmLeggedRobotState2024, wasserman2024legolas}. 
These methods require the network to implicitly learn contact-dependent variations in kinematic reliability without an explicit architectural mechanism for distinguishing and reweighting individual measurements.
However, this implicit learning process can make the estimator overfit to the contact patterns observed during training, leading to degraded performance under unseen locomotion or terrain conditions.

In this work, we aim to improve generalization across diverse gaits and terrains not encountered during training.
To achieve this goal, the proposed architecture is designed to reweight sensor measurements according to their contact-dependent reliability.
For example, model-based contact-aided estimators typically incorporate leg-kinematic measurements only for legs identified as being in stationary contact, whereas measurements from non-contact legs are excluded from the stationary-contact update.

To incorporate contact-dependent measurement weighting into the network architecture, inertial and leg-wise kinematic measurements are represented as individual tokens and processed using an attention mechanism.
Furthermore, the proposed framework adaptively reweights sensor measurements according to the current contact condition without requiring an explicit contact estimation module.
Through this design, the network captures the structural advantages of contact-aided methods, improves generalization compared to conventional learning-based approaches, and at the same time retains the robustness characteristic of learning-based models relative to model-based methods.

These attention-weighted representations serve as inputs to a gated recurrent unit (GRU) \cite{chung2014empirical}, a type of recurrent neural network (RNN).
The RNN updates its hidden representation recursively using the previous hidden state and the current input, inducing a Markovian recurrent structure over the sequence \cite{battaglia2018relational}.
Following \cite{barfoot2024state}, our network incorporates a recursive structure to refine state estimates by capturing the temporal dependencies of robot dynamics. 

The uncertainty associated with the predicted body-frame linear velocity can vary with the contact condition.
For example, during swing phase or significant foot slip, leg-kinematic measurements cannot provide a direct constraint on the body linear velocity through the stationary-contact assumption~\cite{bloesch2013state, hartleyContactaidedInvariantExtended2020}.
In such cases, the network relies more on IMU measurements and temporal motion patterns learned from the training data, as the information provided by leg-kinematic measurements becomes limited.
Consequently, when the motion patterns differ from those observed during training, the error in body linear velocity estimation may increase.
Motivated by this observation, the network is designed to jointly predict the body-frame linear velocity and its associated uncertainty, which enables more reliable integration of the learned pseudo-measurements within the Kalman filtering framework.

The invariant extended Kalman filter (IEKF) \cite{barrauInvariantExtendedKalman2017} is the second component of the proposed system.
In the proposed framework, the IEKF measurement model does not rely on explicit contact estimation or forward-kinematic measurements.
Instead, the network predicts the body-frame linear velocity from proprioceptive inputs, including IMU and joint encoder measurements.
Following the right-invariant body-frame velocity measurement formulation of \cite{youmLeggedRobotState2024}, we incorporate this prediction into the IEKF correction step as a pseudo-measurement.

\section{Attention-Based State Estimation Network}

Building on the system design introduced in Sec.~\ref {sec:system_design}, this section describes the network architecture of the proposed estimator.
The input to the network consists of
$\boldsymbol{\omega}_t$, $\mathbf{a}_t$, $\mathbf{q}_t$, $\dot{\mathbf{q}}_t$, and $\mathbf{e}_{\mathbf{q},t} \triangleq \mathbf{q}^{des}_{t}-\mathbf{q}_t$
, which correspond to the body angular velocity, body linear acceleration, joint position, joint velocity, and joint position tracking error, respectively.
The network predicts the body linear velocity
${}^{B}\mathbf{v} \in \mathbb{R}^3$, 
and the uncertainty associated with the estimate, 
$\mathbf{u} \triangleq [ u_{x}, u_{y}, u_{z} ] \in \mathbb{R}^3$. 
In the remainder of this paper, quantities estimated by the network are denoted by $\bar{(\cdot)}$. 

 
\subsection{Inertial-Leg Tokenization}
\label{subsec:ik_tokenize}

In this section, we describe how each sensor measurement of a legged robot is individually tokenized. 
The input features can be categorized into two groups: inertial measurements ($\boldsymbol{\omega}_t$ and $\mathbf{a}_t$) and leg-related measurements ($\mathbf{q}_t$, $\dot{\mathbf{q}}_t$, and $\mathbf{e}_{\mathbf{q},t}$), which are defined separately for each leg.
Consequently, we construct separate tokens for each inertial measurement and for each leg.
We refer to this scheme as IL-tokenization. 
An example of this tokenization process is illustrated in Fig.~\ref{fig:framework}. 
The proposed network uses three embedding layers in total: two separate embedding layers for the inertial tokens and one shared embedding layer for all leg tokens.
This design accounts for the different physical characteristics of gyroscope and accelerometer measurements, while exploiting the shared measurement structure of the leg tokens across all legs.


In a legged robot, there exist $N_c$ contact points, resulting in $N_c+2$ tokens, consisting of two inertial tokens and $N_c$ leg tokens.
To preserve the ordering of these tokens, we assign a positional index along the IL-axis.
Each embedded token also accumulates information over a history of $N_t$ time steps before being passed to the attention module. 
The temporal ordering across these steps is defined along the history-axis. 
Consequently, positional information along both the IL-axis and the history-axis is encoded using a 2D positional embedding~\cite{devlinBERTPretrainingDeep2019}, formulated as
\begin{align}
\mathrm{PE}(c,t) = \mathbf{e}_c[c] + \mathbf{e}_t[t],
\quad
\mathbf{e}_c \in \mathbb{R}^{(N_c+2) \times D}, \quad
\mathbf{e}_t \in \mathbb{R}^{N_t \times D},
\end{align}
where $c \in \{1,\ldots,N_c+2\}$ denotes the token index along the IL-axis, $t \in \{1,\ldots,N_t\}$ denotes the temporal index along the history-axis, and $D$ is the embedding dimension.
Here, $\mathbf{e}_c$ and $\mathbf{e}_t$ denote learnable embeddings for the IL-axis and history-axis, respectively.


Additionally, inspired by the masked sensory modeling approach of  \cite{liuMaskedSensoryTemporalAttention2025}, we apply random masking.
During training, IL tokens are randomly deactivated to regularize the network and encourage robust token representations.

\subsection{Network Architecture}  


To evaluate the relative importance of the tokenized measurements while maintaining computational efficiency, we adopt the encoder cross-attention module of Perceiver IO~\cite{jaeglePerceiverIOGeneral2022}.
Given $N$ input tokens, the encoder introduces $M$ learnable latent tokens, where $M < N$.
In the cross-attention module, the latent tokens are projected into queries, $\mathbf{Q} \in \mathbb{R}^{M \times D}$, while the input tokens are projected into keys and values, $\mathbf{K}, \mathbf{V} \in \mathbb{R}^{N \times D}$, where $D$ denotes the attention feature dimension.
The computational complexity of self-attention over the input tokens is $O(N^2D)$, whereas that of the encoder cross-attention in Perceiver IO is $O(MND)$.
This structure enables the network to efficiently integrate information from inertial and leg tokens and to learn the relative importance of each measurement under the current state.

Finally, the concatenation of the encoder output and the inertial token at the current time step is used as input to two separate GRU--MLP modules.
The GRU--MLP modules incorporate the temporal ordering of the measurements and estimate the body linear velocity, ${}^{B}\mathbf{v}$, and the uncertainty, $\mathbf{u}$.




\subsection{Loss Function}




A two-stage loss strategy is adopted during training to improve training stability and enable uncertainty learning.
In the initial stage, the network is trained using the mean absolute error loss, $\mathcal{L}_{\mathrm{MAE}}$.
In the subsequent stage, the Gaussian maximum likelihood loss, $\mathcal{L}_{\mathrm{ML}}$, is used to train both the velocity prediction and the associated uncertainty.
Let the velocity prediction residual be $\mathbf{r}_i = {}^{B}\mathbf{v}_i - {}^{B}\bar{\mathbf{v}}_i$.
The losses are then defined as
\begin{align}
\mathcal{L}_{\mathrm{MAE}}
&= \frac{1}{n} \sum_{i=1}^{n} \left\| \mathbf{r}_i \right\|_1 ,
\\
\mathcal{L}_{\mathrm{ML}}
&= \frac{1}{n} \sum_{i=1}^{n}
\left(
\frac{1}{2} \log \det(\bar{\Sigma}_i)
+ \frac{1}{2} \left\| \mathbf{r}_i \right\|_{\bar{\Sigma}_i}^{2}
\right).
\end{align}


Here, 
$n$ denotes the number of training samples. 
$\bar{\Sigma}_i$ is a 3×3 covariance matrix for the $i$-th sample, and the Gaussian maximum likelihood loss and covariance parameterization follow the formulation in \cite{liu2020tlio}. Specifically, $\bar{\Sigma}_i$ is computed from the uncertainty estimate $\bar{u}_i$ predicted by the network as

\begin{align}
\label{eq: uncertainty}
\bar{\Sigma}_i(\bar{u}_i)=\mathrm{diag}(e^{2\bar{u}_{xi}},\,e^{2\bar{u}_{yi}},\,e^{2\bar{u}_{zi}}).
\end{align}




\subsection{Training Details}
\label{sec:Network_TrainingDetails}


The training data are collected in the RaiSim simulator \cite{hwangbo2018per} using a simulated Unitree Go1 quadruped robot.
Trajectories are generated using the DynaFlow locomotion controller ~\cite{leeDynaFlowDynamicsembeddedFlow2025a}, 
and the training set includes only trot locomotion data, excluding other gait patterns such as bound, pace, and pronk.
Each training iteration collects a \SI{2}{\second} trajectory from each simulated environment.
To include near-stationary motions, $10~\%$ of the rollouts are generated by applying random PD targets while the robot remains approximately stationary, instead of using the locomotion controller.
The simulator runs at 1~kHz, and the estimator inputs and labels are recorded at 500~Hz.
The input tokens are accumulated over a history of $N_t=7$ time steps before being passed to the attention module. 
During training, the friction coefficient of the ground to the robot is sampled from $\mathcal{U}(0.3, 1.0)$. 
We also apply random masking to encoder input tokens with a probability of $20~\%$.

During data collection, zero-mean Gaussian noise is added to the simulated IMU measurements, while the kinematic measurements are kept noise-free.
This design is motivated by the fact that IMU measurements are used both for filter propagation and as inputs to the network, which are then used as measurements in the filter update.
As discussed in \cite{liu2020tlio}, such reuse of IMU data introduces correlation between the propagation and update steps, violating the independence assumption of Kalman filtering and potentially leading to estimation inconsistency.
Following previous work \cite{liu2020tlio}, we inject noise only into the IMU measurements during training, thereby reducing the effect of shared IMU errors on the network-predicted measurements.

The proposed network uses a multi-head attention module with two learnable latent tokens, two attention heads, and one attention layer, where each attention head has a dimension of 32. The GRU has a hidden-state dimension of 64. The MLP consists of a single layer with a fully connected layer size of 64.
The network is trained with the Adam optimizer~\cite{kingma2014adam} using a learning-rate schedule with linear warmup for the first 100 iterations up to $5\cdot10^{-4}$, followed by linear decay. 
Among 1000 total iterations, $\mathcal{L}_{\mathrm{MAE}}$ is used for the first 400 iterations, after which $\mathcal{L}_{\mathrm{ML}}$ is used to jointly learn the prediction error and its uncertainty.

\section{Invariant Extended Kalman Filter}
\label{sec:method_inekf}



Following Youm et al.~\cite{youmLeggedRobotState2024}, our system uses the body-frame linear velocity predicted by the network as a pseudo-measurement in the IEKF update.
Unlike their method, however, the proposed method does not use the right-invariant leg kinematics measurement model and therefore does not require explicit contact estimation.



In the IEKF, the state $\mathbf{X}_t \in \mathrm{SE}_{2}(3)$ contains the body orientation, linear velocity, and position, defined as
\begin{align}
\mathbf{X}_t \triangleq
\begin{bmatrix}
\mathbf{R}_t & \mathbf{v}_t & \mathbf{p}_t \\
\mathbf{0}_{1,3} & 1 & 0 \\
\mathbf{0}_{1,3} & 0 & 1 
\end{bmatrix}
\end{align}
where $\mathbf{R}_t \in \mathrm{SO}(3)$ and $\mathbf{v}_t, \mathbf{p}_t \in \mathbb{R}^3$ denote the body orientation,
linear velocity, and position expressed in the world frame, respectively.
Although omitted here for brevity, IMU biases are included in the implementation and propagated following \cite{hartleyContactaidedInvariantExtended2020}.

\subsection{System Dynamics}

The gyroscope and accelerometer outputs are modeled as the true angular velocity and specific force
corrupted by additive zero-mean Gaussian noise:
\begin{equation}
\begin{aligned}
\tilde{\boldsymbol{\omega}}_t &= \boldsymbol{\omega}_t + \mathbf{w}_t^{\boldsymbol{\omega}}, \qquad
\mathbf{w}_t^{\boldsymbol{\omega}} \sim \mathcal{N}(\mathbf{0}, \boldsymbol{\Sigma}^{\boldsymbol{\omega}}) \\ 
\tilde{\mathbf{a}}_t &= \mathbf{a}_t + \mathbf{w}_t^{\mathbf{a}}, \qquad
\mathbf{w}_t^{\mathbf{a}} \sim \mathcal{N}(\mathbf{0}, \boldsymbol{\Sigma}^{\mathbf{a}})
\label{eq:imu_noise_model_no_contact}
\end{aligned}
\end{equation}
where $\mathcal{N}(\cdot)$ denotes a Gaussian distribution, and $\boldsymbol{\Sigma}^{\omega}$ and $\boldsymbol{\Sigma}^{a}$ are the noise covariance matrices of the gyroscope and accelerometer measurements, respectively.

Given these inputs, the nominal state evolves as
\begin{equation}
\dot{\mathbf{R}}_t = \mathbf{R}_t\!\left(\tilde{\boldsymbol{\omega}}_t - \mathbf{w}_t^{\omega}\right)^{\wedge}, \quad
\dot{\mathbf{v}}_t = \mathbf{R}_t\!\left(\tilde{\mathbf{a}}_t - \mathbf{w}_t^{a}\right) + \mathbf{g}, \quad
\dot{\mathbf{p}}_t = \mathbf{v}_t
\label{eq:propagation_no_contact}
\end{equation}
where $\dot{(\cdot)}$ denotes the time derivative, and $\tilde{(\cdot)}$ indicates a measured quantity.
The operator $(\cdot)^{\wedge}:\mathbb{R}^3\!\rightarrow\!\mathfrak{so}(3)$ maps a vector to its corresponding $3\times3$ skew-symmetric matrix.
Finally, $\mathbf{g}\in\mathbb{R}^3$ denotes the gravitational acceleration expressed in the world frame.


\subsection{Right-Invariant Neural Measurement Model}

In the proposed framework, the network-predicted body linear velocity is used as the sole pseudo-measurement for the IEKF correction step. 
Unlike contact-aided IEKF methods~\cite{ hartleyContactaidedInvariantExtended2020} that directly update the filter using leg kinematics under contact assumptions, the proposed measurement model does not require an explicit zero-velocity contact update.
The predicted velocity is formulated as a right-invariant observation, and its associated uncertainty is used to construct the measurement covariance. 
This section presents the resulting neural measurement model and describes how it is incorporated into the IEKF.



We use the body linear velocity predicted by the network as a pseudo-measurement in the IEKF and model its uncertainty with additive zero-mean Gaussian noise:
\begin{equation}
{}^{B}\bar{\mathbf{v}}_t = \mathbf{R}_t^{\top}\mathbf{v}_t + \mathbf{w}_t^{B\mathbf{v}},
\qquad
\mathbf{w}_t^{B\mathbf{v}} \sim \mathcal{N}\!\left(\mathbf{0}_{3,1}, \boldsymbol{\Sigma}^{B\mathbf{v}}\right)
\end{equation}
where $\mathbf{w}_t^{B\mathbf{v}}$ is modeled as zero-mean Gaussian noise representing the uncertainty of the pseudo-measurement.
$\boldsymbol{\Sigma}^{B\mathbf{v}}$ denotes the covariance reconstructed from the uncertainty estimated by the network, as defined in~(\ref{eq: uncertainty}). 

The pseudo-measurement ${}^B\bar{\mathbf{v}}_t$ satisfies the right-invariant observation form \cite{barrauInvariantExtendedKalman2017} and is written as

\begin{equation}
\underbrace{
\begin{bmatrix}
{}^{B}\bar{\mathbf{v}}\\
-1\\
0
\end{bmatrix}
}_{\mathbf{Y}_t}
=
\underbrace{
\begin{bmatrix}
\mathbf{R}_t^{\top} & -\mathbf{R}_t^{\top}\mathbf{v}_t & -\mathbf{R}_t^{\top}\mathbf{p}_t \\
\mathbf{0}_{1,3} & 1 & 0 \\
\mathbf{0}_{1,3} & 0 & 1 \\
\end{bmatrix}
}_{\mathbf{X}_t^{-1}}
\underbrace{
\begin{bmatrix}
\mathbf{0}_{3,1}\\
-1\\
0
\end{bmatrix}
}_{\mathbf{b}}
+
\underbrace{
\begin{bmatrix}
\mathbf{w}_t^{B\mathbf{v}}\\
0\\
0
\end{bmatrix}
}_{\mathbf{V}_t}.
\end{equation}


The Kalman update follows the IEKF formulation of \cite{barrauInvariantExtendedKalman2017}. 
Using the network-predicted body linear velocity as the measurement, the state and covariance update are written as
\begin{equation}
\begin{aligned}
\hat{\mathbf{X}}_t^{+} &= \exp\!\big(\mathbf{K}_t\,\boldsymbol{\Pi}\,\hat{\mathbf{X}}_t\,\mathbf{Y}_t\big)\,\hat{\mathbf{X}}_t, \\
\mathbf{P}_t^{+} &= (\mathbf{I}-\mathbf{K}_t\mathbf{H}_t)\,\mathbf{P}_t\,(\mathbf{I}-\mathbf{K}_t\mathbf{H}_t)^{\top}
+ \mathbf{K}_t\,\hat{\mathbf{N}}_t\,\mathbf{K}_t^{\top},
\end{aligned}
\label{eq:iekf_update}
\end{equation}
where $(\cdot)^+$ denotes the posterior estimate, $\mathbf{K}_t$ is the Kalman gain, $\mathbf{H}_t$ is the measurement Jacobian, and $\hat{\mathbf{N}}_t$ is the measurement noise covariance. 
The gain and covariance terms are computed using the standard IEKF equations in \cite{barrauInvariantExtendedKalman2017}, with
$\boldsymbol{\Pi} \triangleq \begin{bmatrix}\mathbf{I} & \mathbf{0}_{3,2}\end{bmatrix}$ and
$\mathbf{H}_t = \begin{bmatrix}\mathbf{0}_{3,3} & \mathbf{I} & \mathbf{0}_{3,3}\end{bmatrix}$.

\section{Experimental Results}


We consider the following four baselines for evaluation.
\begin{itemize}

\item{\textbf{Proposed without IL tokenization (w/o IL token)}}
To isolate the effect of tokenization independent of the architecture, we adopt the sensory embedding design proposed in \cite{liuMaskedSensoryTemporalAttention2025}, which was originally developed for a locomotion controller rather than state estimation. Specifically, we construct five modality-specific embedding layers for $(\boldsymbol{\omega}_t,\mathbf{a}_t,\mathbf{q}_t,\dot{\mathbf{q}}_t$, and $\mathbf{e}_{\mathbf{q},t})$, and apply positional embeddings along the temporal and sensor axes. 
The resulting tokens are processed by the same Perceiver-IO backbone and prediction heads as in our model, with all training settings kept identical.

\item{\textbf{Neural measurement network (NMN)}}
To evaluate the effect of the attention architecture against a recent learning-based state estimation method, we compare the proposed method with NMN~\cite{youmLeggedRobotState2024}.
Although NMN predicts both contact probabilities and body linear velocity, we use only the body linear velocity predicted by the network to construct the pseudo-measurement, in order to isolate the effect of contact estimation.

\item{\textbf{Contact-aided invariant extended Kalman filter (IEKF)}}
For comparison with model-based methods, we use the contact-aided invariant extended Kalman filter proposed in \cite{hartleyContactaidedInvariantExtended2020}. Contact states are estimated from ground reaction forces (GRF) and used to construct leg kinematic measurements. 

\item{\textbf{Contact-aided invariant extended Kalman filter with slip rejection (IEKF w/ SR)}}
To account for slip in the model-based estimator, we additionally consider a contact-aided IEKF augmented with the slip rejection method proposed in~\cite{kimLeggedRobotState2021}. 



\end{itemize}

\begin{figure}[ht]
    \centering
    \includegraphics[width=\columnwidth]{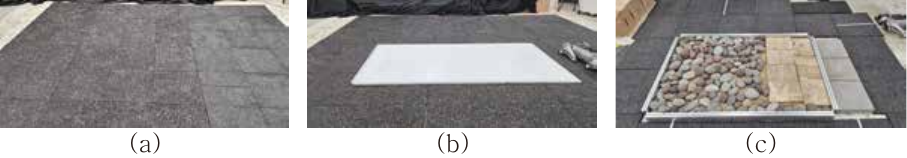}
    \caption{
    \textbf{Experimental environments for each scenario}: 
    (a) rubber surface, 
    (b) slippery surface with boric acid powder on an acetal plate, and 
    (c) debris terrain with pebbles and unsecured wooden planks. 
    The debris terrain was not modeled in the training simulation.
    } 
    \label{fig:exp_envs}
\end{figure}

The experimental environments are shown in Fig.~\ref{fig:exp_envs}. 
Each scenario was conducted three times to examine the consistency of the experimental results. 
The circular scenario combines rubber and slippery surfaces, with the robot traversing each surface for half of the circular path.
We define a slip event as a time step in which the contact estimator used in the IEKF classifies a foot as in contact, while the corresponding ground-truth foot velocity exceeds $0.5~\mathrm{m/s}$ \cite{kimLeggedRobotState2021, youmLeggedRobotState2024}.

Joint position, velocity, and IMU measurements are collected from the Unitree Go1 robot at 500 Hz.
Ground-truth data are acquired at 200 Hz using 10 Vicon Vero V2.2 cameras and interpolated to 500 Hz for evaluation.
The estimator state is initialized from the ground truth, and the IMU bias is set to zero.
All filtering methods use the same covariance parameters for IMU propagation, the prior state, and IMU biases, while method-specific measurement covariances are used only for the corresponding measurement models.

The proposed network requires about 0.12 MFLOPs per inference step, based on the network parameters described in Sec.~\ref{sec:Network_TrainingDetails}.
This is approximately 2.8 times lower than the computational cost of a self-attention mechanism applied to the same input tokens and 3.3 times lower than the network proposed in~\cite{youmLeggedRobotState2024}. 
We evaluated the inference latency using an AMD Ryzen 7 5800X 8-Core Processor over 10,000 samples. 
The network achieved an average inference time of $\mathrm{0.70\pm0.07}$ ms per step.
These results demonstrate that the proposed framework is capable of providing estimated states at a rate sufficient for the high-frequency control loops of legged robots.
Training the network required approximately 1.2 h on an NVIDIA GeForce RTX 3070 GPU.

\subsection{Body Linear Velocity Prediction Results}
\label{sec:network_results}


Before analyzing the performance of the IEKF system, we first evaluate the body-frame linear velocity predicted by the network.
In particular, we examine whether the predicted pseudo-measurements remain reliable under contact conditions different from those observed during training.
We first compare the body-frame linear velocity prediction error in the rubber, slippery, debris, and circular scenarios.
Table~\ref{tab:terrain_network_eval} reports the RMSE over the full trajectory and over slip-event intervals.
Over the full trajectory, the proposed method achieves lower prediction error in most scenarios, with a clear improvement on the debris terrain, where unmodeled contact geometry and movable objects induce contact conditions different from those observed during training.

\begin{table}[t]
\centering
\caption{
Body-frame linear velocity estimation error [m/s] across terrain conditions. 
Each entry reports $\mu$ ($\sigma$).
``Full'' and ``Slip'' denote the full trajectory and slip-event intervals, respectively. 
The symbol $^\dagger$ denotes a terrain condition not modeled in the simulation.
}
\label{tab:terrain_network_eval}
\begin{threeparttable}
\renewcommand{\arraystretch}{1.15}
\setlength{\tabcolsep}{3.5pt}
{\fontsize{7pt}{8pt}\selectfont
\begin{tabular}{l c c c c c}
\Xhline{1.0pt}
\specialrule{0pt}{0.7pt}{0.7pt}
\multicolumn{1}{c|}{Terrain} & \multicolumn{1}{c|}{Split} & Proposed & w/o IL token & NMN & IEKF\\
\specialrule{0pt}{0.7pt}{0.7pt}
\Xhline{1.0pt}
\specialrule{0pt}{0.7pt}{0.7pt}

\multicolumn{1}{c|}{Rub.}
& \multicolumn{1}{l|}{Full}
& \textbf{0.098 (0.001)} & \underline{0.103 (0.002)} & 0.126 (0.003) & 0.128 (0.003)\\
\multicolumn{1}{c|}{}
& \multicolumn{1}{l|}{Slip}
& \textbf{0.102 (0.001)}  & \underline{0.108 (0.002)} & 0.138 (0.003) & 0.141 (0.003)\\
\specialrule{0pt}{0.7pt}{0.7pt}
\hline
\specialrule{0pt}{0.7pt}{0.7pt}

\multicolumn{1}{c|}{Slip.}
& \multicolumn{1}{l|}{Full}
& \underline{0.187 (0.009)} & \textbf{0.185 (0.009)} & 0.192 (0.008) & 0.236 (0.014) \\
\multicolumn{1}{c|}{}
& \multicolumn{1}{l|}{Slip}
& \underline{0.217 (0.011)} & \textbf{0.211 (0.011)} & 0.222 (0.010) & 0.286 (0.020) \\
\specialrule{0pt}{0.7pt}{0.7pt}
\hline
\specialrule{0pt}{0.7pt}{0.7pt}

\multicolumn{1}{c|}{Debr.$^\dagger$}
& \multicolumn{1}{l|}{Full}
& \textbf{0.187 (0.010)} & 0.203 (0.013)  & \underline{0.194 (0.011)} & 0.235 (0.016) \\
\multicolumn{1}{c|}{}
& \multicolumn{1}{l|}{Slip}
& \textbf{0.209 (0.012)} & 0.222 (0.013)  & \underline{0.217 (0.012)} & 0.275 (0.017) \\
\specialrule{0pt}{0.7pt}{0.7pt}
\hline
\specialrule{0pt}{0.7pt}{0.7pt}

\multicolumn{1}{c|}{Circ.}
& \multicolumn{1}{l|}{Full}
& \textbf{0.146 (0.004)} & \underline{0.151 (0.005)} & 0.154 (0.004) & 0.177 (0.007) \\
\multicolumn{1}{c|}{}
& \multicolumn{1}{l|}{Slip}
& \textbf{0.162 (0.006)} & \underline{0.166 (0.006)} & 0.170 (0.005) & 0.204 (0.010) \\
\specialrule{0pt}{0.7pt}{0.7pt}
\Xhline{1.0pt}
\end{tabular}}

\end{threeparttable}
\end{table}
\begin{table}[t]
\centering
\caption{Body-frame linear velocity estimation error [m/s] across different gait patterns. 
Each entry reports $\mu$ ($\sigma$). The symbol $^\dagger$ denotes an unseen gait pattern not included during training.}
\label{tab:gait_network_eval}
\begin{threeparttable}
\renewcommand{\arraystretch}{1.15}
{\fontsize{7pt}{9pt}\selectfont
\begin{tabular}{l c c c c}
\Xhline{1.0pt}
    \multicolumn{1}{c|}{Gait} & Proposed & w/o IL token & NMN & IEKF \\
\Xhline{1.0pt}

\specialrule{0pt}{0.7pt}{0.7pt}
\multicolumn{1}{c|}{Trot}
& \textbf{0.1052 (0.046)} & \underline{0.1135 (0.050)} & 0.1353 (0.061) & 0.1399 (0.065) \\
\hline

\multicolumn{1}{c|}{Bound$^\dagger$}
& \textbf{0.3181 (0.154)} & 0.3485 (0.170) & 0.4013 (0.215) & \underline{0.3401 (0.156)} \\
\hline

\multicolumn{1}{c|}{Pace$^\dagger$}
& \textbf{0.3181 (0.016)} & 0.5339 (0.042) & 0.6766 (0.075) & \underline{0.4245 (0.042)}\\
\hline

\multicolumn{1}{c|}{Pronk$^\dagger$}
& \textbf{0.3629 (0.017)} & 0.4339 (0.027) & 0.5656 (0.070) & \underline{0.4217 (0.071)}\\
\specialrule{0pt}{0.7pt}{0.7pt}

\Xhline{1.0pt}
\end{tabular}}

\end{threeparttable}
\end{table}



\begin{figure}[t] 
    \centering 
    \subfloat[]{ 
        \includegraphics[
            width=\columnwidth,
            trim=0cm 0.1cm 0cm 0.4cm,  
            clip
        ]
        {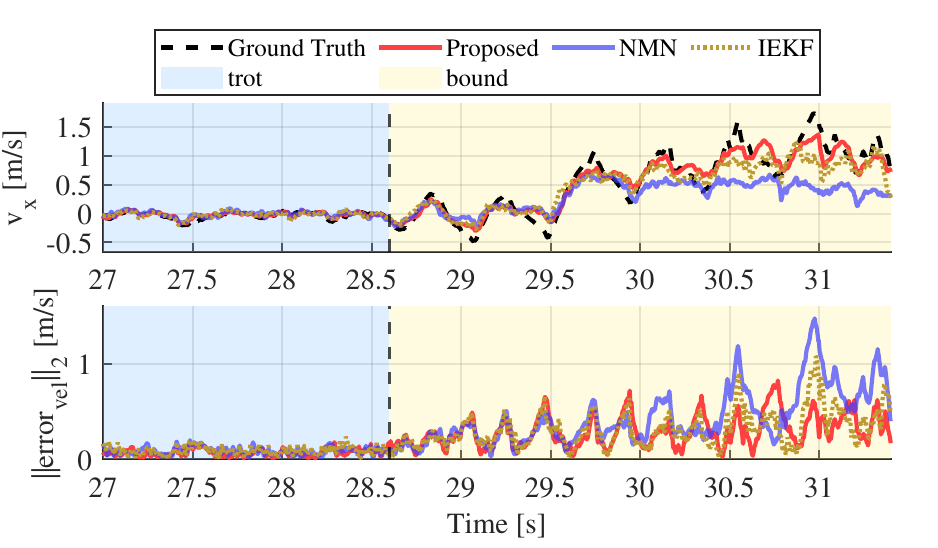} \label{fig:gait_transition_vb_plot} 
        } 
        \vspace{0.01cm} 
    \subfloat[]{ 
        \includegraphics[
            width=\columnwidth,
            trim=0cm 0cm 0cm 0.05cm,  
            clip
        ]
        {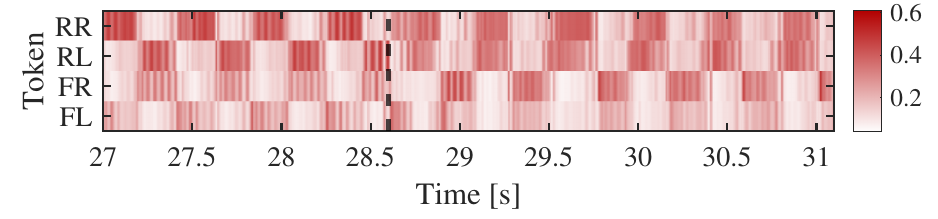} \label{fig:gait_transition_attn_plot} 
        }
    \caption{\textbf{Velocity estimation and leg-token attention during the trot-to-bound transition.}
    The dashed vertical line indicates the transition to bound, which was not included in the training data.
    (a) The upper plot shows the estimated body-frame $x$-velocity, and the lower plot shows the L2 norm of the velocity estimation error.
    NMN exhibits a marked increase in error after the transition, whereas the proposed method maintains lower error under the unseen gait.
    (b) The attention weights of the leg tokens increase around contact events, and their temporal pattern changes with the gait transition.}
    \label{fig:gait_transition_plot}
\end{figure}

On slippery terrain, the proposed method remains competitive, although the variant w/o IL token performs marginally better. 
This small gap may be attributed to extended slip intervals during which foot movement makes it difficult to infer body linear velocity from leg kinematics. 
In such conditions, adaptively balancing inertial and whole-body joint information may become more important than explicitly separating the leg-wise kinematic structure, which is consistent with the role of attention in the proposed architecture. 
These results indicate that the proposed architecture is effective for body-frame linear velocity estimation across diverse terrains.


We further evaluate the prediction performance under gait-induced contact-pattern changes. 
Although the network is trained only on trot data, it is evaluated on bound, pace, and pronk gaits, which are not included in the training dataset. 
Fig.~\ref{fig:gait_transition_vb_plot} shows the prediction result during a trot-to-bound transition, and Table~\ref{tab:gait_network_eval} reports the error across the collected gait data. 
The proposed method maintains a lower prediction error than the learning-based baselines under these unseen gait patterns. 
In particular, the other learning-based baselines show larger errors than the IEKF on unseen gaits, providing a clear contrast under contact-pattern changes.

Table~\ref{tab:gait_network_eval} presents the performance of networks trained only on trot data and evaluated on both trot and unseen gaits. 
We further evaluate the networks in two additional settings to determine whether the proposed method maintains its performance advantage under unseen contact conditions when trained on different source gaits.
First, the networks are trained without trot data and evaluated on trot trajectories.
Second, the networks are trained using pace and pronk data and evaluated on trot and bound trajectories, which are excluded from training.
For these evaluations, we use the same dataset as in Table~\ref{tab:gait_network_eval}. The results are aggregated with equal weighting across gaits so that differences in the number of samples for each gait do not bias the reported mean and standard deviation.
As shown in Table~\ref{tab:gait_ablation}, the proposed network achieves lower prediction error than the learning-based baselines in both evaluation settings.

These results show that the proposed network predicts body-frame linear velocity more reliably than the baselines across terrain-induced and gait-induced contact variations. The effect of IL tokenization and the corresponding attention behavior are analyzed in Sec.~\ref {subsec:il_token_ablation}.

\begin{table}[t]
\caption{
Body-frame linear velocity prediction errors [m/s] under additional gait training configurations.
}
\label{tab:gait_ablation}
\begin{threeparttable}
\renewcommand{\arraystretch}{1.15}
\setlength{\tabcolsep}{5.5pt}
{\fontsize{7pt}{9pt}\selectfont
\begin{tabular}{llccc}
\Xhline{1.0pt}
\multicolumn{1}{c|}{Training gaits} & \multicolumn{1}{c|}{Test gaits} & Proposed & w/o IL token & NMN \\
\Xhline{1.0pt}
\specialrule{0pt}{0.7pt}{0.7pt}
\multicolumn{1}{c|}{Bound/Pace/Pronk} & \multicolumn{1}{c|}{Trot} & \textbf{0.120 (0.05)} & \underline{0.240 (0.09) }& 0.274 (0.13) \\
\hline
\multicolumn{1}{c|}{Pace/Pronk} & \multicolumn{1}{c|}{Trot/Bound} & \textbf{0.222 (0.10)} & \underline{0.320 (0.10)} & 0.332 (0.11) \\
\specialrule{0pt}{0.7pt}{0.7pt}
\Xhline{1.0pt}
\end{tabular}
}
\end{threeparttable}
\end{table}
\begin{table*}[!t]
\centering
\caption{Experimental results. Each entry reports $\mu$ ($\sigma$). The percentage indicates the relative increase in the mean error with respect to the lowest error in each row. The symbol $^\dagger$ denotes a terrain condition not modeled in the simulation.}
\label{tab:quad_terrain_ate_result}
\begin{threeparttable}
\renewcommand{\arraystretch}{1.15}
{\fontsize{7pt}{8pt}\selectfont
\setlength{\tabcolsep}{4.5pt}
\begin{tabular}{l|l|c|c|c|c|c}
\Xhline{1.0pt}
\multicolumn{1}{c|}{Metric} & \multicolumn{1}{c|}{Terrain} & \multicolumn{1}{c|}{IEKF} & \multicolumn{1}{c|}{IEKF w/ SR} & \multicolumn{1}{c|}{NMN} & \multicolumn{1}{c|}{Proposed (w/o IL token)} & \multicolumn{1}{c}{Proposed} \\
\Xhline{1.0pt}
\specialrule{0pt}{0.7pt}{0.7pt}

\multirow{5}{*}{ATE$_{\mathrm{vel}}$ [m/s]}
& Rubber
& 0.1161 (0.054) (+16.5\%)
& 0.1125 (0.052) (+12.9\%)
& 0.1155 (0.051) (+15.9\%)
& {0.1036} (0.044) (+3.9\%)
& \textbf{0.0997 (0.041)} \\
& Slippery
& 0.2036 (0.120) (+32.4\%)
& 0.1840 (0.104) (+19.7\%)
& 0.1687 (0.092) (+9.8\%)
& {0.1582} (0.087) (+2.9\%)
& \textbf{0.1537 (0.088)} \\
& Debris$^\dagger$
& 0.1986 (0.125) (+25.3\%)
& 0.1739 (0.108) (+9.8\%)
& {0.1646} (0.103) (+3.9\%)
& 0.1718 (0.116) (+8.4\%)
& \textbf{0.1584 (0.105)} \\
& Circular
& 0.1562 (0.083) (+22.5\%)
& 0.1435 (0.074) (+12.4\%)
& 0.1393 (0.066) (+9.0\%)
& {0.1372} (0.066) (+7.4\%)
& \textbf{0.1277 (0.065)} \\
\cline{2-7}
& Mean
& 0.1666 (0.104) (+24.7\%)
& 0.1519 (0.090) (+13.7\%)
& 0.1458 (0.082) (+9.1\%)
& \underline{0.1414} (0.085) (+5.8\%)
& \textbf{0.1336 (0.080)} \\
\specialrule{0pt}{0.7pt}{0.7pt}
\hline
\specialrule{0pt}{0.7pt}{0.7pt}

\multirow{5}{*}{ATE$_{\mathrm{rot}}$ [rad]}
& Rubber
& 0.1493 (0.1003) (+11.2\%)
& 0.1708 (0.1159) (+27.2\%)
& \textbf{0.1343 (0.0800)}
& {0.1361} (0.0862) (+1.4\%)
& 0.1362 (0.0863) (+1.4\%) \\
& Slippery
& 0.0366 (0.0319) (+121.2\%)
& 0.0254 (0.0128) (+53.5\%)
& \textbf{0.0163 (0.0067)}
& {0.0166} (0.0073) (+1.4\%)
& 0.0183 (0.0087) (+11.8\%) \\
& Debris$^\dagger$
& 0.0324 (0.0243) (+131.2\%)
& 0.0156 (0.0076) (+11.2\%)
& 0.0162 (0.0090) (+15.7\%)
& \textbf{0.0139 (0.0072)}
& {0.0141} (0.0073) (+0.9\%) \\
& Circular
& 0.2831 (0.2236) (+105.5\%)
& 0.1606 (0.1119) (+16.6\%)
& 0.1471 (0.0961) (+8.4\%)
& \textbf{0.1357 (0.0843)}
& {0.1382} (0.0865) (+1.8\%) \\
\cline{2-7}
& Mean
& 0.1344 (0.1675) (+65.6\%)
& 0.0986 (0.1108) (+21.5\%)
& 0.0835 (0.0904) (+4.0\%)
& \textbf{0.0803 (0.0869)}
& \underline{0.0815} (0.0877) (+1.4\%) \\
\specialrule{0pt}{0.7pt}{0.7pt}
\hline
\specialrule{0pt}{0.7pt}{0.7pt}

\multirow{5}{*}{RE$_{\mathrm{pos}}$ [m]}
& Rubber
& 0.1985 (0.080) (+74.0\%)
& 0.2087 (0.049) (+83.0\%)
& 0.3044 (0.066) (+166.7\%)
& {0.1981} (0.044) (+73.6\%)
& \textbf{0.1141 (0.043)} \\
& Slippery
& 0.2670 (0.127) (+29.3\%)
& 0.3547 (0.107) (+71.7\%)
& {0.2275} (0.067) (+10.1\%)
& 0.3546 (0.094) (+71.7\%)
& \textbf{0.2066 (0.103)} \\
& Debris$^\dagger$
& 0.4247 (0.135) (+226.1\%)
& 0.3381 (0.136) (+159.7\%)
& {0.1517} (0.073) (+16.5\%)
& 0.2795 (0.136) (+114.6\%)
& \textbf{0.1302 (0.070)} \\
& Circular
& {0.2190} (0.089) (+5.9\%)
& 0.2540 (0.087) (+22.8\%)
& 0.2451 (0.067) (+18.5\%)
& 0.3190 (0.072) (+54.3\%)
& \textbf{0.2068 (0.076)} \\
\cline{2-7}
& Mean
& 0.2722 (0.139) (+64.7\%)
& 0.2847 (0.115) (+72.2\%)
& \underline{0.2347} (0.087) (+42.0\%)
& 0.2870 (0.108) (+73.6\%)
& \textbf{0.1653 (0.086)} \\
\specialrule{0pt}{0.7pt}{0.7pt}
\hline
\specialrule{0pt}{0.7pt}{0.7pt}

\multirow{5}{*}{RE$_{\mathrm{vel}}$ [m/s]}
& Rubber
& 0.1773 (0.078) (+18.9\%)
& 0.1705 (0.071) (+14.4\%)
& 0.1569 (0.065) (+5.2\%)
& \textbf{0.1474 (0.058)}
& {0.1503} (0.059) (+2.0\%) \\
& Slippery
& 0.3068 (0.159) (+43.9\%)
& 0.2646 (0.134) (+24.1\%)
& 0.2507 (0.125) (+17.5\%)
& \textbf{0.2133 (0.110)}
& {0.2245} (0.116) (+5.3\%) \\
& Debris$^\dagger$
& 0.2875 (0.152) (+21.7\%)
& 0.2552 (0.133) (+8.0\%)
& {0.2465} (0.130) (+4.3\%)
& 0.2519 (0.140) (+6.6\%)
& \textbf{0.2363 (0.132)} \\
& Circular
& 0.2315 (0.111) (+23.7\%)
& 0.2030 (0.095) (+8.5\%)
& 0.1987 (0.092) (+6.2\%)
& {0.1970} (0.087) (+5.2\%)
& \textbf{0.1893 (0.087)} \\
\cline{2-7}
& Mean
& 0.2477 (0.136) (+25.0\%)
& 0.2207 (0.116) (+11.4\%)
& 0.2109 (0.112) (+6.4\%)
& \underline{0.2007} (0.108) (+1.3\%)
& \textbf{0.1982 (0.106)} \\
\specialrule{0pt}{0.7pt}{0.7pt}
\hline
\specialrule{0pt}{0.7pt}{0.7pt}

\multirow{5}{*}{RE$_{\mathrm{rot}}$ [rad]}
& Rubber
& 0.0334 (0.0160) (+5.8\%)
& 0.0357 (0.0149) (+13.1\%)
& 0.0425 (0.0136) (+36.7\%)
& \textbf{0.0311 (0.0124)}
& {0.0316} (0.0125) (+1.7\%) \\
& Slippery
& 0.0213 (0.0138) (+61.8\%)
& 0.0148 (0.0075) (+12.0\%)
& \textbf{0.0132 (0.0070)}
& {0.0132} (0.0069) (+0.1\%)
& 0.0133 (0.0070) (+0.9\%) \\
& Debris$^\dagger$
& 0.0198 (0.0105) (+38.7\%)
& 0.0152 (0.0083) (+6.4\%)
& {0.0142} (0.0081) (+0.0\%)
& \textbf{0.0143 (0.0082)}
& 0.0144 (0.0082) (+1.5\%) \\
& Circular
& 0.0574 (0.0284) (+47.7\%)
& 0.0423 (0.0203) (+8.9\%)
& 0.0433 (0.0216) (+11.5\%)
& {0.0394} (0.0202) (+1.1\%)
& \textbf{0.0391 (0.0200)} \\
\cline{2-7}
& Mean
& 0.0344 (0.0246) (+34.8\%)
& 0.0281 (0.0188) (+10.2\%)
& 0.0295 (0.0205) (+15.8\%)
& \textbf{0.0255 (0.0176)}
& \underline{0.0256} (0.0175) (+0.4\%) \\
\specialrule{0pt}{0.7pt}{0.7pt}
\Xhline{1.0pt}

\end{tabular}

}
\end{threeparttable}
\end{table*}

\subsection{State Estimation Results with IEKF System}


To evaluate the overall effectiveness of the proposed estimation framework, we compare it with both model-based and learning-based baselines across different terrain scenarios.
The quantitative results are reported in Table~\ref{tab:quad_terrain_ate_result}.
The definitions of ATE and RE follow~\cite{zhang2018tutorial}, and RE is computed over 5-second sub-trajectories. 
The error units are $\mathrm{rad}$ for rotation, $\mathrm{m/s}$ for velocity, and $\mathrm{m}$ for position.


Compared with IEKF w/ SR, which mitigates slip effects by increasing the contact-related covariance when slip is detected~\cite{kimLeggedRobotState2021}, the proposed method achieves lower estimation errors. 
Unlike slip rejection, the proposed method does not explicitly detect slip, but instead uses the body-frame linear velocity predicted by the network as a pseudo-measurement.
These results indicate that the learned pseudo-measurement can improve estimation performance in scenarios where contact model violations occur frequently.

When NMN and the proposed method are compared, the proposed method reduces errors in ATE$_\mathrm{vel}$, RE$_\mathrm{pos}$, and RE$_\mathrm{vel}$ in several terrains. 
This trend is consistent with the network-level results in Table~\ref{tab:terrain_network_eval}, where the proposed architecture yields lower body linear velocity prediction errors. 
These gains are also reflected in the final estimation results, suggesting that the inductive bias introduced in the proposed architecture helps the learned measurement generalize more robustly to real-world conditions.

Proposed (w/o IL token), which modifies only the tokenization scheme of the proposed architecture, shows a 73.6\% larger RE$_\mathrm{pos}$ on average than the proposed method.
This indicates that the proposed IL tokenization reduces position drift in the final state estimate.
This difference is particularly noticeable in the debris terrain.
In this scenario, movable pebbles and unsecured objects frequently induce unstable contacts, and we observed that the locomotion controller deviates from the behavior observed in simulation.
Under these conditions, Proposed (w/o IL token) produces a $114.6 \%$ larger $\mathrm{RE}_\mathrm{pos}$ than the proposed method.
These results suggest that IL tokenization helps the network use leg-wise information more effectively under contact conditions that are not well represented in simulation.


\subsection{Ablation Study on IL Tokenization}
\label{subsec:il_token_ablation}

\begin{figure}[t]
    \centering
    \includegraphics[width=\columnwidth]{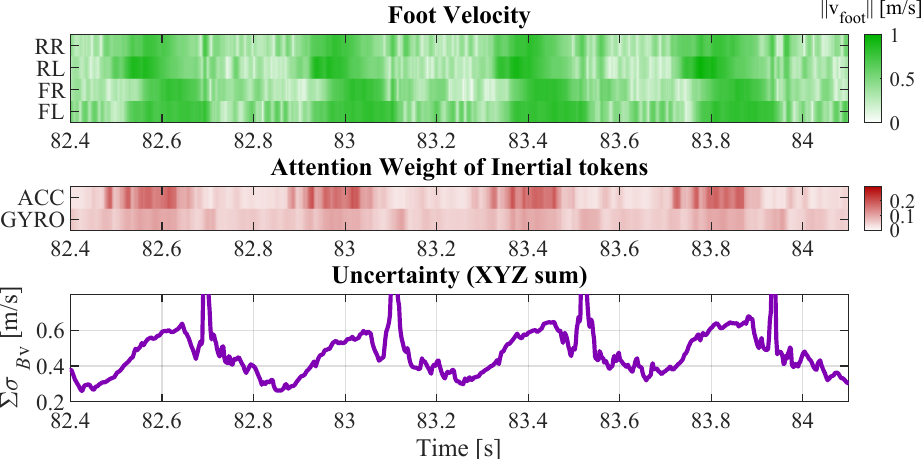}
    \caption{\textbf{Inertial-token attention and uncertainty during pronk gait.}
    Flight phases are identified by simultaneously high foot velocities across all four feet, indicating the absence of stationary contact constraints.
    During these intervals, inertial-token attention increases, particularly the attention weight assigned to the accelerometer token, alongside an increase in the predicted body-frame velocity uncertainty.
    This result is consistent with the contact-aided model-based analysis that body linear velocity becomes unobservable when stationary contacts are unavailable~\cite{bloeschStateEstimationLegged2013, hartleyContactaidedInvariantExtended2020}.}
    \label{fig:pronk_uncertainty}
\end{figure}

As discussed in Sec.~\ref{sec:network_results}, the proposed network shows lower prediction error than the variant without IL tokenization under several contact conditions different from the training distribution.
IL tokenization also makes the attention distribution more interpretable by separating inertial and leg-wise kinematic information into distinct tokens.
We therefore examine how the attention weights assigned to these tokens change under different contact patterns.

Fig.~\ref{fig:gait_transition_attn_plot} provides an interpretable explanation for this behavior. 
The figure visualizes the attention weights during the trot-to-bound transition, summed over the history axis for each leg token. 
During trot, the attention weights increase jointly for the RR--FL and RL--FR pairs, whereas during bound, they increase jointly for the RR--RL and FR--FL pairs. 
This suggests that the network adjusts its attention to leg tokens associated with the current contact condition, rather than relying on a fixed gait-specific contact pattern.
This behavior is consistent with contact-aided state estimation, where leg kinematics are used as reliable measurements when the feet are assumed to be in stationary contact.

We further analyze the pronk gait to examine how the network responds when the stationary contact constraints become weak.
In pronk, all four feet make and break contact nearly simultaneously, and stationary contact constraints are unavailable during the flight phase.
As shown in Fig.~\ref{fig:pronk_uncertainty}, intervals where the velocities of all four feet increase simultaneously can be interpreted as flight phases.
During these intervals, the attention weights assigned to the inertial tokens increase, especially for the accelerometer token.
The predicted uncertainty of the body-frame linear velocity also increases in the same intervals.
This behavior is consistent with the contact-aided state estimation view that body-frame linear velocity becomes unobservable from leg kinematics in the absence of stationary contact.
\section{CONCLUSIONS}


We presented an attention-based proprioceptive state estimation framework designed to improve generalization by allowing the network to reweight each measurement according to the current contact condition.
The proposed network was trained in simulation using only trotting data generated by a locomotion controller. 
Experimental results showed that the proposed framework maintained robust estimation performance on terrain conditions that are difficult to model in simulation and under gait changes that produced unseen contact patterns. 
Moreover, using only the pseudo-measurement predicted by the network, the proposed framework outperformed the conventional contact-aided model-based estimator. 
The attention weight analysis further provided insight into how the network reweights inertial and leg measurements under unseen contact conditions. 
These results indicate that model-based structural priors can be effectively incorporated into neural architectures for proprioceptive state estimation in legged robots.




\bibliographystyle{IEEEtran}
\bibliography{IEEEabrv, references}

\vfill

\end{document}